%% file: main.tex
\documentclass[runningheads]{llncs}

\usepackage{eccv}

\usepackage{eccvabbrv}

\usepackage{graphicx}
\usepackage{booktabs}

\usepackage[accsupp]{axessibility}  

\usepackage[pagebackref,breaklinks,colorlinks,citecolor=eccvblue]{hyperref}

\usepackage{orcidlink}
\usepackage[utf8]{inputenc} 
\usepackage[T1]{fontenc}    
\usepackage{url}            
\usepackage{amsfonts}       
\usepackage{nicefrac}       
\usepackage{microtype}      
\usepackage{xcolor}         
\usepackage{caption}    

\usepackage{multirow}
\usepackage{bbding}
\usepackage{pifont}
\usepackage{extpfeil}
\usepackage{makecell}
\usepackage[table, dvipsnames]{xcolor}
\usepackage{wrapfig}
\usepackage{enumitem}

\usepackage{subcaption}

\definecolor{mygray}{RGB}{210, 210, 210}
\definecolor{mygrayL}{RGB}{240, 240, 240}

\newcommand{\bluetext}[1]{\colorbox[HTML]{D4E6F1}{\textbf{#1}}}

\newcommand{\greentext}[1]{\colorbox[HTML]{E6F8E0}{#1}}

\usepackage{tabularx} 
\begin{document}

\title{ShaRP: SHAllow-LayeR Pruning for Efficient Video Large Language Models}

\titlerunning{ShaRP for Efficient Video Large Language Models}

\author{
Yingjie Xia\inst{1}$^\ast$ \and
Tao Liu\inst{1}$^\ast$ \and
Jinglei Shi\inst{1}$^\dagger$ \and
Qingsong Xie\inst{3} \and
Heng Guo\inst{4} \and
Jian Yang\inst{1} \and
Xi Wang\inst{2}
}

\authorrunning{Y. Xia et al.}

\institute{
VCIP \& TMCC \& DISSec, College of Computer Science, Nankai University
\and
LIX, Ecole Polytechnique, IP Paris
\and
OPPO Research Institute
\and
Beijing University of Posts and Telecommunications
}

\maketitle
{
\renewcommand{\thefootnote}{\fnsymbol{footnote}}
\footnotetext[1]{Equal Contribution. \quad$^{\dagger}$ Corresponding author.}
}
\input{sec/0_abstract}

\input{sec/1_intro}

\input{sec/2_related_work}

\input{sec/3_method}

\input{sec/4_experiment}

\input{sec/5_conclusion}

\bibliographystyle{unsrt}
\bibliography{main}
\end{document}

%% file: sec/0_abstract.tex
\begin{abstract}

Video Large Language Models (VLLMs) incur substantial prefilling cost due to the large number of visual tokens. While attention-based token pruning offers a promising acceleration strategy, applying it at shallow decoder layers often causes severe performance degradation under high compression ratios, limiting its practical benefits.
In this work, we uncover an overlooked failure mode in shallow-layer attention pruning: attention scores in early decoder layers can become unreliable indicators of token utility, resulting in unstable token selection under aggressive compression. We show that this effect arises from the joint influence of insufficient token interaction, content-agnostic positional bias, and redundancy among high-attention tokens, which together distort attention-based importance estimation before informative representations fully emerge.
Motivated by this insight, we propose \textbf{ShaRP}, a unified pruning framework that restores reliable attention-based token selection by jointly improving local information aggregation, calibrating positional bias, and reducing redundancy.
Extensive evaluations show that ShaRP preserves about 97.2\% of the original performance while reducing TFLOPs by 86\% and achieving a 5.1$\times$ speedup in the prefilling stage, providing a scalable solution for efficient VLLM inference, with no training.
\keywords{Efficient Inference \and Visual Token Pruning \and Attention Bias}

\end{abstract}

%% file: sec/1_intro.tex
\section{Introduction}
\label{sec:intro}
Vision Large Language Models (VLLMs)~\cite{lin2023video,lillava,zhang2024video,wang2024qwen2,bai2025qwen2,xu2024pllava} have brought deep video understanding task into a new era. 
They use a vision encoder to convert video streams into many tokens, which interact during pre-filling.
Such a process thus incurs quadratic computational complexity relative to the token count. 
To mitigate this, numerous 
studies~\cite{chen2024image,xing2024pyramiddrop,sun2025adatp,endo2025feather,shang2025llava,yang2025visionzip,bolya2022token,zhang2025beyond,tao2025dycoke,yin2025lifting,huang2024prunevid,liu2025video,chen2025variation,hyun2025multi,shen2025fastvid,shao2025holitom,ma2025mmg,liless,lv2026hippo,fanflashvid} have focused on reducing visual token counts to compress information and accelerate VLLM inference.

\begin{figure}[!t]
    \centering
    \includegraphics[width=1\linewidth]{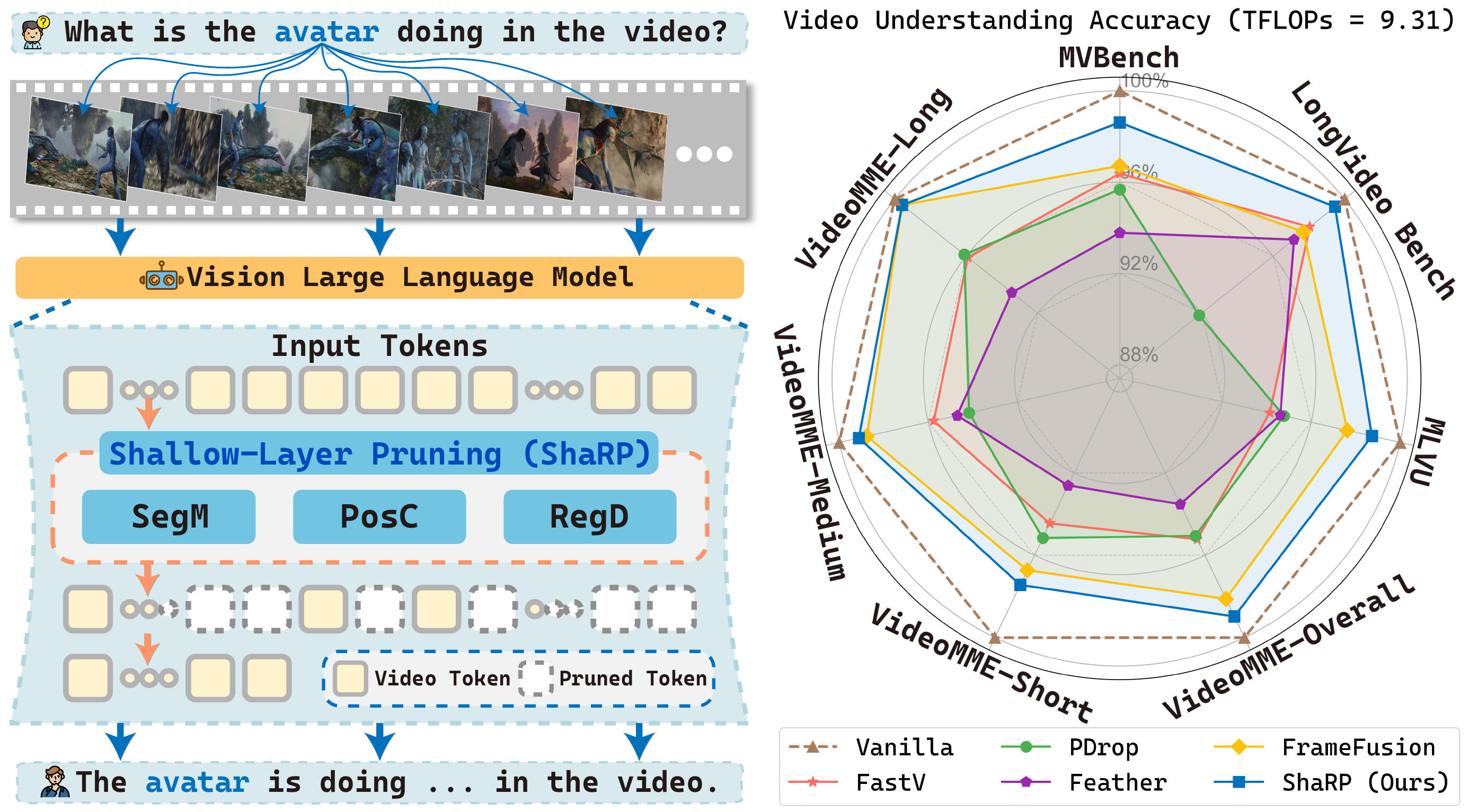} 
    \caption{
\textbf{Left:} ShaRP is a training-free, attention-based framework for inner-LMM token pruning in VLLMs.  \textbf{Right:} ShaRP delivers superior performance across video understanding benchmarks.
}
    \label{fig:teaser}
\end{figure}

\begin{figure}[!t]
    \centering
    \includegraphics[width=1\linewidth]{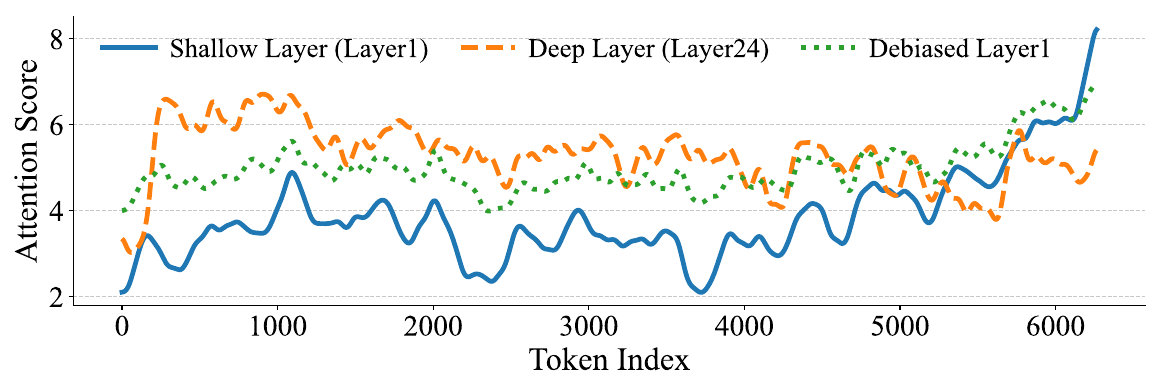} 
    \caption{\textbf{Layer-wise Attention Scores.}
Shallow-layer attention exhibits strong positional bias, assigning higher scores to later tokens, whereas deep-layer attention is semantically guided. After debiasing, shallow-layer attention closely aligns with that of deep layers, enabling early and reliable pruning.}
    \label{fig:intro}
\end{figure}

Among various compression strategies, attention-based ones~\cite{chen2024image,xing2024pyramiddrop,sun2025adatp,endo2025feather,lin2025boosting,yin2025lifting} offer a compelling approach.
Unlike visual cue-guided pruning that operates purely on visual similarity~\cite{yang2025visionzip,alvar2025divprune,sun2025llava,fu2025framefusion,hyun2025multi,tao2025dycoke,fanflashvid}, attention-based methods leverage the inherent cross-modal reasoning ability of the model to estimate token importance, inherently aligning the pruning process with semantic relevance.
It not only enhances interpretability~\cite{chefer2021transformer,zheng2024attention,rigotti2021attention} but also enables adaptive compression that dynamically focuses on content most relevant to the textual query.
Representative studies such as FastV~\cite{chen2024image} show that visual information becomes less critical in deeper decoder layers, motivating aggressive pruning early. However, this strategy struggles with high compression on long videos, as performance drops sharply, leading subsequent works~\cite{xing2024pyramiddrop,endo2025feather,shao2025holitom} to prune more conservatively in deeper layers, losing much of the computational savings. 
Viewed differently, we advocate the potential of attention in shallow layers as an effective criterion for token selection.
To explore this, we conduct a pilot study comparing attention scores across shallow and deep layers. As shown in Fig.~\ref{fig:intro}, visual analysis reveals that visual tokens near the sequence end receive abnormally high attention in shallow layers. When properly tackled, shallow layer attention exhibits distributions similar to deep ones, enabling early yet reliable pruning.

In our paper, we contend that early-stage pruning failure is not inevitable; it can be achieved at an early stage (Fig.~\ref{fig:teaser}) by addressing \textbf{three overlooked factors}:

\noindent
(1) \emph{Information remains under-interacted in shallow layers}:
As revealed in HiMAP~\cite{yin2025lifting} and visiPurner~\cite{fan2025mathcal}, shallow layers mainly facilitate cross-modal exchange, while interactions among same-modality visual tokens remain underdeveloped.
Once pruning at shallow layers, some retained visual tokens often exhibit high-norm anomalies~\cite{darcet2023vision}; they receive high attention scores yet carry low information density (i.e., attention collapse in~\cite{tang2025seeing,zou2025don}). 
This frequently occurs with semantically irrelevant tokens, such as background visual tokens. 
This issue arises because the softmax function assigns non-zero attention to all tokens, causing many low-relevance or non-priority tokens to receive unnecessary attention.

\noindent
(2) \emph{Content-agnostic positional biased} can be introduced~\cite{liu2025global}: 
Such insufficient interaction, as discussed in (1), leaves shallow-layer attention overly dependent on positional priors.
Taking RoPE~\cite{su2024roformer} as an example, due to its long-term decay property, the model assigns unreasonably high scores to tokens near the end of the sequence (i.e., to later video frames). In methods like FastV~\cite{chen2024image}, when the compression ratio increases (e.g., pruning over 50\% tokens), most retained tokens come from the end of the sequence. This inherent bias in attention scores makes attention-based pruning in shallow layers unreliable.

\noindent(3) \textit{Post-Pruning stage faces an importance-redundancy dilemma}:
Even after mitigating positional bias through methods like (2), relying solely on attention as an importance indicator can still cause the model to repeatedly retain high-attention~\cite{darcet2023vision} tokens from adjacent frames or similar backgrounds. 
As observed in the first layer of LLaVA-Video-7B~\cite{zhang2024video}, over half of the tokens selected based on attention scores exhibit high redundancy~\cite{fu2025framefusion}, with cosine similarity exceeding 0.9 to other selected tokens. 
Under a constrained compression rate, these similar register tokens~\cite{darcet2023vision} "crowd out" those that should preserve the holistic context of the scene, ultimately leading to degraded performance under high compression ratios.

To this end, we propose \textbf{ShaRP}, a training-free shallow-layer pruning framework built on a single insight: \emph{in early decoder layers, raw cross-attention is a miscalibrated importance signal}. ShaRP restores a \emph{selective} importance ranking that favors informative and non-redundant tokens via a lightweight three-step calibration loop.
First, to \textit{enhance information interaction}, inspired by~\cite{li2025attention}, which observed that locally focused heads in Transformers form near-diagonal block attention, segmenting complex reasoning into phrasal units. We adapt this to the visual domain by partitioning videos into content-aware segments, intensifying local interactions and alleviating attention collapse~\cite{tang2025seeing,zou2025don}.
Second, our method \textit{calibrates positional bias} in attention scores through a rapid global bias estimation.
Finally, to \emph{balance between the importance and redundancy}, high-attention register tokens are refined through a three-step pipeline composed of pre-filtering, deduplication, and post-filling, which eliminates redundancies while enriching the representation with broader contextual information.
Extensive experiments demonstrate that our approach achieves high compression in shallow LLM layers while maintaining state-of-the-art performance, establishing a new paradigm for attention-based token pruning.

Our contributions are summarized as follows:
\begin{itemize}
\item \textbf{Revealing the Factors of Early Pruning Failure:}
We show that attention allocation collapse, PE bias, and register token redundancy are three factors that limit attention-based pruning performance in shallow decoder layers. 
\item \textbf{A Simple yet Effective Training-Free Framework:}
We introduce a \emph{plug-and-play} framework, ShaRP, integrating segment-aware causal masking, positional bias calibration and register token deduplication to effectively address the above factors and improve compression performance.
\item \textbf{Competitive Performance Against SOTAs:}
Comprehensive experiments demonstrate that ShaRP achieves superior performance compared to other token compression approaches across different VLLM backbones, e.g., LLaVA-OneVision~\cite{lillava} and Qwen2.5-VL~\cite{bai2025qwen2}, across multiple benchmarks.
\end{itemize}

%% file: sec/2_related_work.tex
\section{Related work}
\label{sec:related}
\subsection{Visual Cue-Guided Token Pruning}

Since their emergence, VLLMs~\cite{team2023gemini,liu2024improved,liu2024llavanext,lillava,gao2023llama,li2023blip,wang2024qwen2} have demonstrated significant potential across diverse applications, including VQA~\cite{goyal2017making,hudson2019gqa,zhang2025teaching}, video captioning~\cite{tang2021clip4caption,chen2024sharegpt4video}, contextual reasoning~\cite{peng2025lmm,team2025kimi,wang2023visionllm} etc.. 
Earlier approaches employ a limited number of learnable queries to aggregate visual information.
For instance, VideoLLaMA~\cite{zhang2023video} adopts a Q-Former~\cite{li2023blip} to distill dense visual features into compact query sets.
While GlimpsePrune~\cite{zeng2025glimpse} introduces a trainable Visual Importance Predictor (VIP) within the VLLM to score and prune visual tokens at different layers. 
The recent method EPIC~\cite{wen2025efficient} adopts a Progressive Consistency Distillation strategy to re-finetune the model, enabling it to learn rich visual information even from incomplete visual inputs.
Token pruning can be achieved in a training-free manner. Among these, LLaVA-PruMerge~\cite{shang2025llava}, VisionZip~\cite{yang2025visionzip}, and VisPruner~\cite{zhang2025beyond} leverage the inherent sensitivity of vision encoders to visual tokens to identify and retain visually dominant tokens. DivPrune~\cite{alvar2025divprune} formulates pruning as a Max-Min Diversity Problem (MMDP), ensuring that the set of retained visual tokens possesses maximum diversity. Nevertheless, the temporal redundancy inherent in video data calls for temporal-aware pruning mechanisms. DyCoke~\cite{tao2025dycoke} merges tokens within fixed temporal windows. 
Other methods~\cite{sun2025llava,hyun2025multi,shen2025fastvid,shao2025holitom} remap the numerous visual tokens back to the frame level, enabling them to identify and merge highly similar visual content across frames while retaining unique content. 
Among these, STTM~\cite{hyun2025multi} performs multi-granular spatio-temporal token merging.
FastVID~\cite{shen2025fastvid} performs segmentation based on video content, combining temporal segmentation with spatio-temporal merging. 
HoliTom~\cite{shao2025holitom} utilizes dynamic programming to merge redundant information across adjacent frames.

\subsection{{Text--Vision Attention-Based Token Pruning}}
As the other family of token pruning approaches, pruning visual tokens based on text–vision attention scores in the transformer layers of the LLM decoder has proven effective for reducing redundancy while preserving task-relevant information. 
For example, FastV~\cite{chen2024image} pioneers this path as it observes strong sparsity in visual-token attention at deeper layers and thus prunes a fixed proportion of tokens based on shallow-layer attention scores.
However, this approach is susceptible to positional bias and attention collapse, which inhibits high compression ratios in shallow layers. 
To achieve smoother compression dynamics, PyramidDrop (PDrop)~\cite{xing2024pyramiddrop} adopts a hierarchical pruning mechanism while Feather~\cite{endo2025feather} introduces a de-RoPE-based attention criterion and performs multi-stage pruning via uniform sampling, yet the bias from the positional encoding at earlier layers and from the visual encoder remains in the de-RoPEed attention scores. 
HoloV~\cite{zou2025don} mitigates attention collapse by pruning tokens within uniform spatial partitions based on contextual diversity and saliency. However, this block-wise strategy may scatter attention across irrelevant regions, weakening focus on question-relevant content in long videos.
HiMAP~\cite{yin2025lifting} adopts a layer-adaptive strategy that prunes tokens with low cross-modal contribution in shallow layers and those less relevant to intra-visual aggregation in deeper layers.
In addition, FrameFusion~\cite{fu2025framefusion} performs layer-wise compression in the shallow layers of LMMs by exploiting the similarity among visual tokens, while applying attention-based importance pruning to the remaining tokens in deeper layers.
However, relying solely on attention scores can overemphasize a few salient tokens, losing holistic context and risking semantic distortion when merging dissimilar tokens. This underscores the need for both reliable attention evaluation and careful token merging to preserve meaningful information.

%% file: sec/3_method.tex
\begin{figure}[!t] 
\centering \includegraphics[width=1\linewidth]{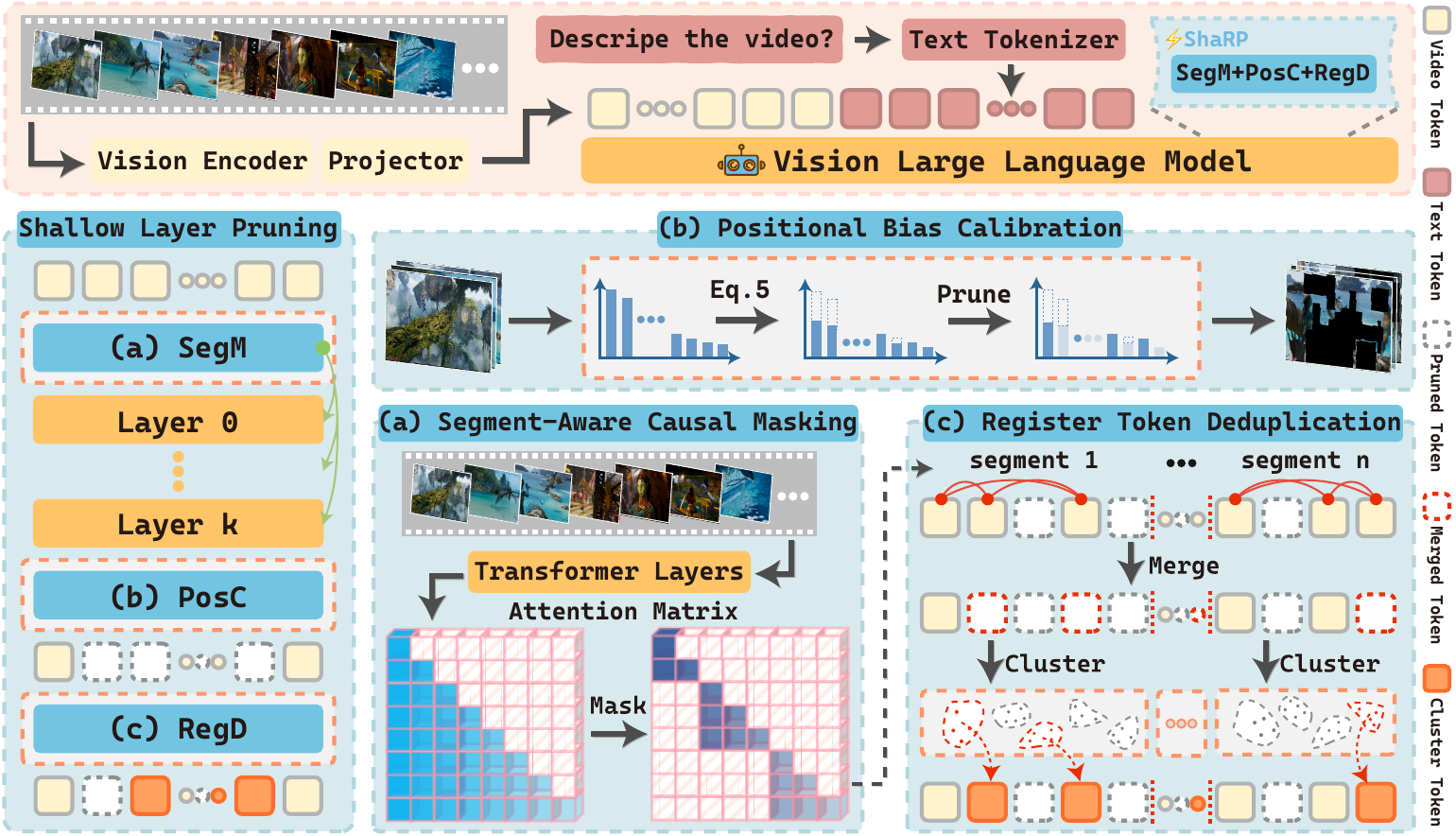} 
\caption{\textbf{Pipeline of ShaRP.}
\textit{ShaRP} performs token pruning in video LLMs through three sequential modules: 
\textbf{(a) Segment-Aware Causal Masking (SegM)} partitions video tokens into content-consistent segments and restricts attention within them; 
\textbf{(b) Positional Bias Calibration (PosC)} debiases shallow-layer attention scores, aligning them with semantically meaningful distributions;  
\textbf{(c) Register Token Deduplication (RegD)} refines pruned tokens by merging redundant ones and replenishing diverse representatives.} 
\label{fig:pipeline} 
\end{figure}

\section{Preliminary Knowledge}
\label{sec:preliminary}

\noindent\textbf{Multimodal Causal Attention:} 
VLLMs such as LLaVA~\cite{lillava} and Qwen~\cite{wang2024qwen2} process both encoded visual and text tokens with causal attention mechanism. Given a visual token set $\mathcal{H}_{v}$
and a textual token set $\mathcal{H}_{t}$,
their tokens are concatenated into $\boldsymbol{H} \in \mathbb{R}^{N \times d}$ with $N$ the length of sequence and $d$ the hidden dimension, and iteratively processed within decoder layer by layer, via the causal attention mechanism.

Specifically, $\boldsymbol{H}$ is first linearly projected into query $\boldsymbol{Q}$, key $\boldsymbol{K}$, and value $\boldsymbol{V}$ matrices, each in $\mathbb{R}^{N \times d}$. The attention score matrix $\boldsymbol{A} \in \mathbb{R}^{N \times N}$ and output $\boldsymbol{O}$ are computed as:
\begin{equation}
\label{eq:self_attention}
\boldsymbol{O} = \text{Softmax}(\boldsymbol{A}) \boldsymbol{V} \quad \text{and} \quad \boldsymbol{A} = \frac{\boldsymbol{Q}\boldsymbol{K}^{\top}}{\sqrt{d}} + \boldsymbol{M},
\end{equation}
where $\boldsymbol{M}$ is a causal mask with entries $m_{i,j} = -\infty$ for all $i < j$ and $i,j$ are token indices, to effectively mask the upper triangular region. This ensures that each token only attends to the past tokens.

\noindent\textbf{Rotary Positional Encoding:} 
RoPE~\cite{su2024roformer} encodes relative position through rotary transformations. Instead of using additive positional biases, 
it applies position-specific rotary matrices $\boldsymbol{R}_{i}, \boldsymbol{R}_{j} \in \mathbb{R}^{d \times d}$ to the query $\boldsymbol{q}_{i}$ and key $\boldsymbol{k}_{j}$ at positions $i$ and $j$. The resulting relative position-aware attention score $\tilde{a}_{i,j}$ based on Eq.~\ref{eq:self_attention} is thus given by:
\begin{equation}
\label{eq:rope}
\tilde{a}_{i,j} = \frac{(\boldsymbol{q}_i\boldsymbol{R}_i ) (\boldsymbol{k}_j\boldsymbol{R}_j )^{\top}}{\sqrt{d}}+m_{i,j} = \frac{\boldsymbol{q}_i \boldsymbol{R}_{j-i} \boldsymbol{k}_j^{\top}}{\sqrt{d}}+m_{i,j},
\end{equation}
where $(j-i)$ denotes relative distance between $\boldsymbol{q}_{i}$ and $\boldsymbol{k}_{j}$.

\section{Methodology}
\label{sec:methodology}

Current attention-based pruning methods struggle in early decoder layers due to attention allocation \textbf{\textit{collapse}}, positional encoding \textbf{\textit{bias}}, 
and register token \textbf{\textit{redundancy}}. As illustrated in Fig.~\ref{fig:pipeline}, we propose a training-free framework that respectively addresses these issues through effective yet simple techniques, detailed as follows:

\label{sec:method}
\subsection{Segment-Aware Causal Masking}
\label{sec:aggregation}

In the shallow layers of large video-language models, high-norm but semantically weak tokens often emerge, typically corresponding to redundant or low-information regions~\cite{darcet2023vision,yang2025visionzip}.
Relying solely on attention scores for importance-based pruning exacerbates this issue, as the causal attention and softmax normalization mechanisms inherently assign non-zero weights to all tokens, allowing even irrelevant ones to receive disproportionately high attention~\cite{darcet2023vision}.
Meanwhile, the insufficient interaction among visual tokens at early layers~\cite{yin2025lifting} further limits the model’s ability to aggregate meaningful representations, leaving these anomalous tokens unrefined.

To mitigate this \emph{attention collapse}~\cite{tang2025seeing,zou2025don} that emerges in shallow layers, 
we draw inspiration from \emph{locally focused heads}~\cite{li2025attention}, which concentrate attention within small, semantically coherent regions.
We thus introduce \textit{segment-aware causal masking} that constrains visual attention within content-consistent video segments.
This localized masking preserves the causal decoding property while enhancing intra-segment visual interaction, ensuring that information flows more effectively within relevant visual contexts.

Precisely, given the visual token set 
$\mathcal{H}_{v}$, 
we first apply average pooling to tokens from the same frame, obtaining a frame-wise averaged token set $\mathcal{H}_{f} = \{\bar{h}^{i}_{f}\}$, where $i$ is the frame index. 
To partition $\mathcal{H}_{v}$ into semantically consistent segments, we insert segment boundaries by evaluating the cosine similarity $sim(\cdot)$~\cite{xia2015learning} between consecutive tokens in $\mathcal{H}_{f}$ against a predefined threshold $\tau_{\text{seg}}$:
\begin{equation}
\label{eq:boundary}
\text{Insert } B_{k} = \begin{cases} \text{yes,} & sim(\bar{h}^{i-1}_{f},\bar{h}^{i}_{f})<\tau_{\text{seg}}
\\ \text{no,} & \text{otherwise} \end{cases},i=2,..,|\mathcal{H}_{f}|,
\end{equation}
where $|\mathcal{H}_{f}|$ is the size of $\mathcal{H}_{f}$. 
The boundaries help us to split $\mathcal{H}_{v}$ into contiguous non-overlapping segments $\mathcal{S} = \{\mathcal{S}^{i}_{v}\}$.
Based on $\mathcal{S}$, we then perform a masked attention computation where visual tokens $\mathcal{H}_{v}$ within the same segment remain mutually visible, while those from different segments are completely isolated. 
Such segment-aware visual mask is thus constructed as a block-diagonal matrix as show in Fig.~\ref{fig:pipeline}(a).
The segment-aware masked attention scores between one visual token $h^{i_{0}}_{v}$ to others follows:
\begin{equation}
    \boldsymbol{A}_{i_{0},j} = [\underbrace{-\infty,...,-\infty}_{\text{outside segment}},\underbrace{\tilde{a}_{i_{0},i_{0}-l_{0}},...,\tilde{a}_{i_{0},i_{0}}}_{\text{inside segment (length $l_{0}$)}},\underbrace{-\infty,...,-\infty}_{\text{causality}}],
\end{equation}
which implies that attentions originally scattered across all segments will be \emph{squeezed} into one single segment. This localized masking effectively narrows the operational range of the softmax, concentrating attention weights within semantically consistent regions (i.e, smaller triangles) and enhancing the information flow of visual feature aggregation.

\begin{figure}[!t]
\centering
\includegraphics[width=1\linewidth]{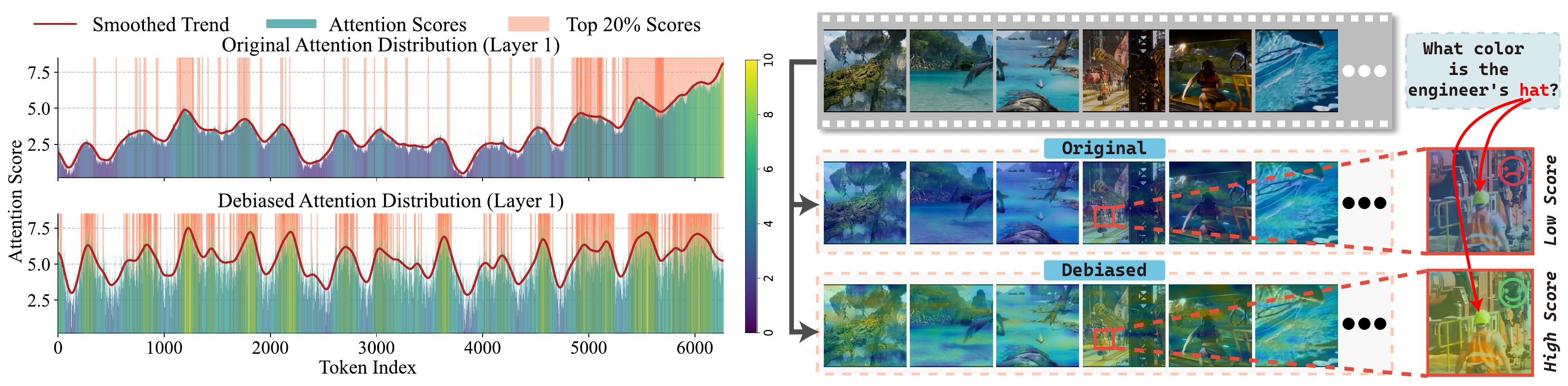}
\caption{\textbf{Original vs. Debiased Attention.}
\textbf{Left:} Averaged attention scores on VideoMME~\cite{fu2025video} at 20\% token retention. After debiasing, attention is more evenly distributed across frames, aligning better with question-relevant regions.
\textbf{Right:} Cross-attention between the last text token and video tokens. In the original setting, semantically important middle frames receive \textbf{low attention scores}, while after debiasing, attention correctly focuses on relevant segments with \textbf{high scores}.}
\label{fig:avg_attn}
\end{figure}

\subsection{Positional Bias Calibration}
\label{sec:debiasing}
Positional Encodings (PE) equip VLLMs with token order information, whereas also introduce inherent positional bias~\cite{barbero2024round} that favors tokens at specific locations. 
A typical example is the long-term decay issue~\cite{tang2025seeing} in RoPE (Eq.~\ref{eq:rope}), where attention scores decrease as the relative token distance $(j-i)$ increases. 
While in the context of video token pruning, existing methods~\cite{chen2024image,xing2024pyramiddrop,endo2025feather} usually identify more important visual tokens $h^{i}_{v}\in \mathcal{H}_{v}$ by referring attention scores $\tilde{a}_{i,T}$ from the last text token $h^{T}_{t}\in \mathcal{H}_{t}$.
They typically assume that visual tokens with higher scores are more relevant to the query. 

However, we argue that positional bias significantly distorts the distribution of attention scores, which may lead to misinterpretation of this mechanism.
As shown in Fig.~\ref{fig:avg_attn} (left-top), our analysis on the VideoMME~\cite{fu2025video} benchmark at 20\% retention reveals that most top-ranked tokens (orange regions) cluster near the final frames, reflecting a clear positional bias rather than true semantic relevance. Fig.~\ref{fig:avg_attn} (right-middle) further illustrates a case where the question targets a mid-video event, yet attention remains skewed toward later frames, leaving semantically important regions under-attended (blue regions).

Therefore, we propose to calibrate PE bias before pruning them with respect to their attention scores. 
Notably, PE techniques span a broad spectrum, ranging from Absolute PEs (APE)~\cite{raffel2020exploring,press2021train,chi2022kerple} to Relative ones (RPE)~\cite{su2024roformer,chowdhery2023palm,touvron2023llama}, with each exhibiting distinct properties (e.g., additivity, rotation invariance) and bias patterns. Furthermore, as VLLMs may employ multiple PE strategies for different modules (e.g., visual encoder, text encoder, decoder), simply removing one~\cite{endo2025feather} or all PEs for debiasing will cause the remaining PE components to dominate or even inference collapse.

Without loss of generality or effectiveness, and inspired by the additive formulation of relative position encoding~\cite{li2023functional}, we propose a \emph{simple, content-agnostic, and PE-type-independent method} to estimate the positional encoding bias inherent in VLLMs. To this end, we input a video of entirely black frames into the VLLM, ensuring all visual tokens are homogeneous and deprived of information. In this setup, any deviation in the attention score $b_{i,T}$ between $h^{i}_{v}$ and $H^{T}_{t}$ directly reflect the bias introduced by positional encodings.
Larger $b_{i,T}$ values indicate stronger bias, requiring greater compensation on $\tilde{a}_{i,T}$, and \emph{vice versa}. To remove this bias, we simply subtract it from the attention scores:
\begin{equation}
    \hat{a}_{i,T} = \tilde{a}_{i,T} -\lambda b_{i,T},
\end{equation}
where $\lambda$ is a hyperparameter that controls the strength of the compensation.
Note that this compensation on $\hat{a}_{i,j}$ is only for token \emph{ranking and pruning}, the value of each token remains unaltered. After applying our proposed debiasing strategy, as shown in Fig.~\ref{fig:avg_attn} (left-bottom), the overall attention distribution becomes more balanced across the temporal dimension. Meanwhile, in the right-bottom panel, attention is more concentrated on semantically relevant regions (e.g., hats), effectively alleviating the positional bias issue.
Importantly, this bias estimation requires only a single lightweight inference with bias-free inputs and is performed once per pruning step, while all subsequent attention computations remain fully compatible with FlashAttention~\cite{dao2022flashattention}.

\subsection{Pruning and Register Token Deduplication}
\label{sec:redundancy}
\noindent\textbf{Pruning.} 
After applying segment-aware causal masking (Sec.~\ref{sec:aggregation}) and attention debiasing (Sec.~\ref{sec:debiasing}), 
following prior works~\cite{chen2024image} we perform token pruning based on the debiased attention scores $\hat{a}_{i,T}$. 
Specifically, we select the top-$K$ visual tokens with the highest scores across the entire sequence. 
These retained tokens are distributed among different segments $\{\mathcal{S}_{v}^{i}\}$, 
and form a compact register token set $\mathcal{H}_{\mathrm{reg}}$.

\noindent\textbf{Register Token Deduplication.} 
Although high-attention tokens are preserved after pruning, many still remain highly similar in both spatial and temporal dimensions, as discussed in Sec.~\ref{sec:intro}. To further compress the representation, we introduce a three-step post-pruning refinement for register tokens \textbf{within each segment}, consisting of pre-filtering, deduplication, and post-filling, taking register tokens within segment $\mathcal{S}^{0}_{v}$ as an example, i.e. $\mathcal{H}^{0}_{reg}\subset\mathcal{S}^{0}_{v}$:

(1) \textbf{Token Pre-Filtering}: Non-register tokens whose similarity to any token in $\mathcal{H}^{0}_{reg}$ exceeds $\tau_{filter}$ are absorbed into their nearest similar register token. The remaining non-register tokens (noted as $\mathcal{H'}^{0}_{nreg}$) are preserved for Step (3).

(2) \textbf{Token Deduplication}: We perform iterative merging on register tokens by sequentially comparing adjacent tokens. A token is absorbed into its preceding token if their similarity exceeds $\tau_{merge}$, otherwise it becomes the new pivot. This continues until the end of the sequence is reached, giving a deduplicated register token set $\mathcal{H'}^{0}_{reg}$ for Step (3).

(3) \textbf{Token Post-Filling}: Step (2) reduces redundancy among register tokens by merging similar ones. To maintain a fixed number of register tokens under the preset compression rate, we cluster the non-register tokens in $\mathcal{H'}^{0}_{nreg}$ and replenish $\mathcal{H'}^{0}_{reg}$ with selected cluster centers. Starting from the first token in $\mathcal{H'}^{0}_{nreg}$, we perform iterative clustering: a subsequent token joins the current cluster $\mathcal{C}_{k}$ if its similarity exceeds $\tau_{cluster}$, otherwise it initiates a new cluster. For the resulting clusters, we compute the diversity score of each of their centers $c_k$ as:
\begin{equation}
Div_{k} = \left[1 - \frac{1}{|\mathcal{H'}^{0}_{reg}|} \sum_{h^{i}_{reg} \in \mathcal{H'}^{0}_{reg}} \text{sim}(c_k, h^{i}_{reg})\right] + \beta |\mathcal{C}_{k}|,
\label{eq:diversity_score}
\end{equation}
where $|\mathcal{H'}^{0}_{reg}|$ and $|\mathcal{C}_{k}|$ denote set sizes and $\beta$ is a balancing hyperparameter. This score considers both the average similarity to register tokens and the cluster size. Centers with the highest diversity are selected as replenished register tokens.
Finally, the filled register token set is forwarded to deeper decoder layers for inference.

\input{table/table1}
\input{table/qwen}

%% file: table/table1.tex
\begin{table*}[t]
\centering
\caption{\textbf{Methods comparison.} Performance of representative token pruning methods on LLaVA-OneVision~\cite{lillava}, grouped by comparable TFLOPs. $R$ denotes the retention ratio of visual tokens for each method. TFLOPs are reported for video tokens(see supplementary materials for more details.). \textbf{Best} results are marked in \bluetext{blue}, and second-best in \greentext{green}.}
\resizebox{\linewidth}{!}
{
\begin{tabular}{c|cc|ccccccc|cc}
\toprule
 \multirow{2}{*}{\textbf{Method}} & \multirow{2}{*}{\makecell{\textbf{Retention}\\\textbf{Ratio $\boldsymbol{R}$ ↓}}} & \multirow{2}{*}{\makecell{\textbf{TFLOPs ↓}}} & \multirow{2}{*}{\textbf{MVBench ↑}} & \multirow{2}{*}{\makecell{\textbf{LongVideo}\\\textbf{Bench ↑}}} & \multirow{2}{*}{\textbf{MLVU ↑}} & \multicolumn{4}{c}{\textbf{VideoMME ↑}}    &\multicolumn{2}{|c}{\multirow{2}{*}{\textbf{Average ↑}}} \\ \cline{7-10}
&   &  &                          &                                 &                       & \textbf{Overall}      & \textbf{Short}       & \textbf{Medium}       & \textbf{Long}     & &               \\
\textbf{Duration}                         &  &                              & \textbf{16s}                      & \textbf{1$\sim$60min}                    & \textbf{3$\sim$120min}         & \textbf{1$\sim$60min} & \textbf{1$\sim$3min} & \textbf{3$\sim$30min} & \textbf{30$\sim$60min} & \textbf{Score}            &  \textbf{\%}          \\ \midrule
\rowcolor{mygrayL}LLaVA-OneVision~\cite{lillava}                 & 100\%  & 40.8         & 57.9 & 56.4   & 63.2  & 58.6         & 70.3   & 56.6    & 48.8
&  59.0       & 100        \\ \midrule

FastV~\cite{chen2024image}$_{\text{ECCV'24}}$                   &  25\%  &11 &55.3 &55.1 &60.3  &56.5   &\greentext{68.2}  &55    &46.3 &56.8         &96.2        \\
PDrop~\cite{xing2024pyramiddrop}$_{\text{CVPR'25}}$               & 28.5\%  &11        &\greentext{56.7}     &54.2     &61.1            &56.4 &68.0         &53.6         &47.7        &57.1   &96.7        \\
Feather~\cite{endo2025feather}$_{\text{ICCV'25}}$              & 25\%&11  &54.8  &54.7 	&60.8 	&55.0 	&65.3 	&53.6 	&46.2 	&56.3 	&95.4        \\
Dycoke~\cite{tao2025dycoke}$_{\text{CVPR'25}}$              & 25\%&8.7  &53.1  &49.5 	&55.8 	&51.0 	&61.1 	&48.6 	&43.2 	&52.4 	&88.7        \\
STTM~\cite{hyun2025multi}$_{\text{ICCV'25}}$              & 21.2\%&10.5  &50.8  &51.1 	&58.7 	&54.9 	&65.2 	&53.5 	&46.1 	&53.9 	&91.3        \\

FrameFusion~\cite{fu2025framefusion}$_{\text{ICCV'25}}$     & 31\%&-     & \greentext{56.7}	&\greentext{55.8} 	&\greentext{62.2}	&\greentext{57.4}	&68.1 	&\greentext{55.4} 	&\greentext{48.6}	& \greentext{57.8}	&\greentext{97.9}        \\

\textbf{ShaRP}     & 28.2\% &11                            &\bluetext{57.1}    & \bluetext{56.9}      &\bluetext{62.6}    & \bluetext{57.7}    &\bluetext{68.4}        &\bluetext{55.7}    &\bluetext{48.9}      &\bluetext{58.6}      &\bluetext{99.3}        \\ \midrule

FastV~\cite{chen2024image}$_{\text{ECCV'24}}$    &  20\%  &9.31  &55.8       &\greentext{55.3} &59.5       &55.8                   & 66.4         &54.2         &46.8                     &56.6         &95.9        \\
PDrop~\cite{xing2024pyramiddrop}$_{\text{CVPR'25}}$               & 19.6\% &9.31    &55.4        &51.8    &59.9     &55.7         &66.9  &53.3    &46.9           &55.7         &94.4        \\
Feather~\cite{endo2025feather}$_{\text{ICCV'25}}$              & 20\%& 9.31                               &54.3 	&54.8 	&59.8 	&54.8 	&65.1 	&53.6 	&\greentext{45.6} 	&55.9 	&94.7        \\
FrameFusion~\cite{fu2025framefusion}$_{\text{ICCV'25}}$  & 26\%   &-	&\greentext{56.0} 	&55.1 	&\greentext{61.7} 	&\greentext{57.5} 	&\greentext{68.0} 	&\greentext{55.9} 	&\bluetext{48.6} 	&\greentext{57.6}   &\greentext{97.5}     \\

\textbf{ShaRP}        & 23.5\% &9.31   &\bluetext{57.1}                      & \bluetext{55.8}     &\bluetext{61.9}               & \bluetext{57.6}         & \bluetext{68.1}        &\bluetext{56.1}          & \bluetext{48.6}          &\bluetext{58.1}         &\bluetext{98.5}        \\ \midrule

FastV~\cite{chen2024image}$_{\text{ECCV'24}}$                   &  15\% &7.66  &54.9   & \greentext{54.8}        & 58.4             &54.7          &65.1         &53.4          &45.7           &55.7         &94.4        \\
PDrop~\cite{xing2024pyramiddrop}$_{\text{CVPR'25}}$               & 11.5\% &7.66 &54.5 &50.0    &59.1        &54.0                   &64.3         &51.6         &46.1         &54.4           &92.2                 \\
Feather~\cite{endo2025feather}$_{\text{ICCV'25}}$        & 15\%&7.66                                &54.5 	&53.0 	&59.1	& 54.7	&65.1 	&53.3 	&45.6 	&55.3 	&93.7        \\
FrameFusion~\cite{fu2025framefusion}$_{\text{ICCV'25}}$              & 21\%&-      &\bluetext{57.0}	&54.3 	&\greentext{60.6}	&\bluetext{56.6}	&\greentext{67.0} 	&\bluetext{55.2}	&\greentext{47.6}	& \greentext{57.1}	&\greentext{96.8}        \\

\textbf{ShaRP}             & 18.8\% & 7.66   &\greentext{56.5}   & \bluetext{55.9}      &\bluetext{62.5}               & \greentext{56.3}         & \bluetext{67.8}        &\greentext{53.5}          & \bluetext{47.7}          &\bluetext{57.8}         &\bluetext{98.0}        \\ \midrule

FastV~\cite{chen2024image}$_{\text{ECCV'24}}$                   &  10\% &6.03  &53.5   & 52.4        & 56.2             &52.6          &61.1         &51.6          &45.2           &53.7         &90.9        \\
PDrop~\cite{xing2024pyramiddrop}$_{\text{CVPR'25}}$               & 1.5\% &6.03    &48.2                      &44.7                             &52.5                   &45.0         &49.6         &45.8         &39.6           &47.6         &80.6        \\
Feather~\cite{endo2025feather}$_{\text{ICCV'25}}$              & 10\%&6.03                               & 53.6	&52.7 	&57.6 	&53.6	&62.4 	&52.7 	&45.6 &54.4 & 92.1       \\
FrameFusion~\cite{fu2025framefusion}$_{\text{ICCV'25}}$              & 16\%&-    & \greentext{55.1}	&\greentext{53.0}	&\greentext{58.3}	&\greentext{55.5} 	& \greentext{65.8}	&\greentext{54.1}	&\greentext{46.7}	&\greentext{55.0}   &\greentext{94.0}    \\

\textbf{ShaRP}             & 14.1\% &6.03  &\bluetext{56.4}       & \bluetext{55.3}                            &\bluetext{62.0}               & \bluetext{55.8}         & \bluetext{65.9}        &\bluetext{54.6}          & \bluetext{46.9}    &\bluetext{57.4}         &\bluetext{97.2}        \\

\bottomrule
\end{tabular}
}
\label{tab:table1}
\vspace{-8pt}
\end{table*}

%% file: table/qwen.tex
\begin{table}[t]
\centering
\caption{\textbf{Cross-backbone method comparison.} 
Performance of representative token pruning methods on LLaVA-Video~\cite{zhang2024video},Qwen2-VL~\cite{wang2024qwen2} and Qwen2.5-VL~\cite{bai2025qwen2}
grouped by comparable TFLOPs, demonstrating strong generalization across model backbones.}
\resizebox{\linewidth}{!}
{
\begin{tabular}{c|cc|cccccc|cc}
\toprule
\multirow{2}{*}{\textbf{Method}}  
& \multirow{2}{*}{\makecell{\textbf{Retention}\\ \textbf{Ratio $\boldsymbol{R}$ ↓} }} 
& \multirow{2}{*}{\makecell{\textbf{TFLOPs ↓}}} 
& \multirow{2}{*}{\textbf{MVBench ↑}} 
& \multirow{2}{*}{\makecell{\textbf{LongVideo}\\\textbf{Bench ↑}}} 
& \multicolumn{4}{c}{\textbf{VideoMME ↑}}    
 &\multicolumn{2}{|c}{\multirow{2}{*}{\textbf{Average ↑}}} \\ 
 \cline{6-9}
&     &   &    &   & \textbf{Overall}  & \textbf{Short}  & \textbf{Medium}   & \textbf{Long}    &   &          \\
 \textbf{Duration}   &  &  & \textbf{16s}  & \textbf{1$\sim$60min}      & \textbf{1$\sim$60min} & \textbf{1$\sim$3min} & \textbf{3$\sim$30min} & \textbf{30$\sim$60min} & \textbf{Score}            &  \textbf{\%}          \\ \midrule

    \rowcolor{mygrayL}LLaVA-Video~\cite{zhang2024video} & 100.0\% & 80.2 & 59.4 & 58.8 & 64.3 & 77.3  & 62.3 & 53.2 & 60.8  &100 \\ \hline
    FastV~\cite{chen2024image}$_{\text{ECCV'24}}$        & 20.0\% & 16.3  & 54.5 & 54.6 & 58.7 & 68.8 & 58.5 & 48.8 & 55.9  &91.9 \\
    PDrop~\cite{xing2024pyramiddrop}$_{\text{CVPR'25}}$        & 19.5\% & 16.3  & 55.1 & 55.0 & 60.3 & 70.6 & \greentext{60.4} & 49.9 & 56.8  &93.4 \\
    
    Fearher~\cite{endo2025feather}$_{\text{ICCV'25}}$      & 20.0\%  & 16.3 & 54.0 & 55.0 & 59.7 & 71.3 & 58.2 & 49.5 & 56.2  & 92.4\\
    Dycoke~\cite{tao2025dycoke}$_{\text{CVPR'25}}$        & 27.3\%  &16.3  & 51.3 & 53.5 & 56.6 &66.3  &54.1  &49.4  &53.8   & 88.5\\

    STTM~\cite{hyun2025multi}$_{\text{ICCV'25}}$        & 17.1\%  &17.0  & 49.2 & 54.5 &59.8 &68.7  &59.8  &50.9  &54.5   & 89.6\\
    
    Framefusion~\cite{fu2025framefusion}$_{\text{ICCV'25}}$      & 25.5\%  & - & \greentext{56.5} & \greentext{57.6} & \greentext{61.3} & \greentext{72.8} & 59.3 & \greentext{51.9}  & \greentext{58.5}  & 96.2\\
    
    \textbf{ShaRP}        & 23.5\%  & 16.3 & \bluetext{58.2} & \bluetext{58.1} & \bluetext{62.7} & \bluetext{75.7} & \bluetext{60.9} & \bluetext{52.5} & \bluetext{59.7} & \bluetext{98.2}  \\ \midrule

\rowcolor{mygrayL}Qwen2-VL~\cite{wang2024qwen2}          & 100\%  & 14.1         & 64.8 & 55.1    & 56.7         & 67.4   & 54.2    & 48.3
&  58.9       & 100        \\ \midrule

FastV~\cite{chen2024image}$_{\text{ECCV'24}}$                  &  10\%  &2.2 &52.2 &45.6 &50.9     &57.4  &48.4    &\bluetext{46.7} &49.6         &84.2        \\

PDrop~\cite{xing2024pyramiddrop}$_{\text{CVPR'25}}$         & 3.4\% &2.2    &46.1                      &45.3                             &46.6                   &52.8         &44.3         &42.6                    &46.0         &78.1        \\
Feather~\cite{endo2025feather}$_{\text{ICCV'25}}$             & 10\%&2.2                    &56.8 	&48.6 	 	&51.1	&60.2	&47.9 	&45.1 &52.2 &88.6        \\
FrameFusion~\cite{fu2025framefusion}$_{\text{ICCV'25}}$             & 16.4\%&-    &\greentext{59.5} 	&\greentext{47.4}	&\greentext{52.0}	 	&\greentext{61.3} 	&\greentext{49.0}	&45.8	&\greentext{53.0}   &\greentext{90.0}    \\

\textbf{ShaRP}            & 13.9\% &2.2  &\bluetext{63.9}       & \bluetext{53.3}                            &\bluetext{52.2}               & \bluetext{61.5}         & \bluetext{49.1}        &\greentext{46.1}          &\bluetext{56.5}             &\bluetext{95.9}        \\ \midrule

    \rowcolor{mygrayL}Qwen2.5-VL~\cite{bai2025qwen2}    & 100.0\% & 22.4 & 67.1 & 58.8 & 62.0 & 73.5 &60.7 &51.7  & 62.6 & 100 \\ \hline
    FastV~\cite{chen2024image}$_{\text{ECCV'24}}$         & 10.0\% & 3.4  & 60.2 & 47.7 & 54.6 & 61.5 & 55.0 & 47.3 & 54.2  &86.6\\

    PDrop~\cite{xing2024pyramiddrop}$_{\text{CVPR'25}}$         & 3.4\% &3.4    &56.0           &47.4                                                &52.4         &54.4         &52.6                    &49.3     &51.9    &83.0        \\
    
    Fearher~\cite{endo2025feather}$_{\text{ICCV'25}}$       & 10.0\%  & 3.4 & 61.6 & 50.5 & 55.0 & 53.3 & 53.9 & 47.8 & 55.7  & 89.0\\
    Framefusion~\cite{fu2025framefusion}$_{\text{ICCV'25}}$      & 16.4\% & -  & \greentext{63.3} & \greentext{50.4} & \greentext{56.4} & \greentext{56.4} & \greentext{66.0} & \greentext{55.1}  & \greentext{56.7} & \greentext{90.6} \\
    \textbf{ShaRP}        & 14.1\% & 3.4  & \bluetext{63.8} & \bluetext{53.9} &\bluetext{56.8}  & \bluetext{66.3} & \bluetext{55.6} & \bluetext{48.3}  & \bluetext{58.2} & \bluetext{93.0} \\

\bottomrule
\end{tabular}
}
\label{tab:qwen}
\end{table}

%% file: sec/4_experiment.tex
\section{Experiments}
\subsection{Experimental Settings}

\noindent\textbf{Benchmarks.} 
We evaluate ShaRP on four widely used video understanding benchmarks: MVBench~\cite{li2024mvbench,shahroudy2016ntu}, LongVideoBench~\cite{wu2024longvideobench}, MLVU~\cite{zhou2024mlvu}, and VideoMME~\cite{fu2025video}. 
These datasets contain videos from 10 seconds to 2 hours, with over 10,000 questions spanning diverse scenarios, providing a solid test of our method's robustness and generality.

\noindent\textbf{Baselines.}
We compare ShaRP with eight representative \emph{training-free} token pruning approaches.
For image compression,
FastV~\cite{chen2024image} and PDrop~\cite{xing2024pyramiddrop} perform attention-guided pruning starting from the 2nd and 4th decoder layers, respectively.
Feather~\cite{endo2025feather} removes positional embeddings at the 2nd layer and applies uniform token sampling before pruning.
VisionZip~\cite{yang2025visionzip} performs outer-LLM frame-level pruning using a vision encoder, upon which we apply ShaRP as an inner-LLM compression stage for additional reduction.
For video token compression,
FrameFusion~\cite{fu2025framefusion} conducts progressive layer-wise pruning based on token similarity and attention importance, we compute an equivalent retention ratio based on the average proportion of retained tokens to ensure comparable computational cost.
DyCoke~\cite{tao2025dycoke} performs temporal merging prior to LLM processing, while STTM~\cite{hyun2025multi} reduces redundancy through spatiotemporal token merging.
HoliTom~\cite{shao2025holitom} adopts a two-stage design with outer- and inner-LLM pruning; we replace its inner pruning stage with ShaRP to evaluate compatibility.
Unless otherwise specified, we follow the official backbone configurations.

\noindent\textbf{Implementation Details.}
We implement ShaRP on four representative multimodal LLMs, i.e., LLaVA-OneVision-7B~\cite{lillava}, LLaVA-Video-7B~\cite{zhang2024video}, Qwen2-VL-7B~\cite{wang2024qwen2}, and Qwen2.5-VL-7B~\cite{bai2025qwen2}, which embody distinct architectural designs to ensure the robustness of our approach and we apply pruning at the first decoder layer. All experiments are conducted on NVIDIA A40 GPUs using the LMMs-Eval framework~\cite{zhang2024lmms}.
The hyperparameters are empirically determined and fixed to $\lambda = 0.6$, $\beta = 0.008$, $\tau_{seg} = 0.9$ for all experiments. See more details in the supplementary materials.

\subsection{Main Results}

\noindent\textbf{Enhanced Overall Performance.}
For LLaVA-OneVision~\cite{lillava}, $32 \times 196$ tokens from 32 sampled frames are evaluated under retention ratios about $R = \{25\%, 20\%, 15\%, 10\%\}$, with FLOPs matched across methods for fair comparison. Tab.~\ref{tab:table1} reports average scores and accuracy under different $R$ and TFLOPs.
\emph{\textbf{FastV}}~\cite{chen2024image} achieves decent performance at higher retention ratios (e.g., $R=25\%$), yet its accuracy declines sharply under aggressive pruning (e.g., $R=10\%$), indicating that raw attention scores are unreliable when few tokens remain.
\emph{\textbf{Feather}}~\cite{endo2025feather} alleviates positional bias by removing local positional encoding (at a certain layer) and uniform sampling. It shows limited performance under low retention ratios as many tokens retained are irrelevant to the query, achieving only 95.4\% of the vanilla model’s performance even at 25\% retention. 
\emph{\textbf{PDrop}}~\cite{xing2024pyramiddrop} employs a progressive, layered pruning method. To achieve comparable FLOPs, this strategy necessitates aggressive pruning in deeper layers, which leads to very low accuracy even if the shallow layers retain the complete set of tokens. 
\emph{\textbf{DyCoke}}~\cite{tao2025dycoke} and \emph{\textbf{STTM}}~\cite{hyun2025multi} reduce redundancy through temporal or spatiotemporal merging; however, they exhibit notable performance degradation even at moderate compression levels. For example, at approximately $25\%$ retention, their average performance drops to 88.7\% and 91.3\%, respectively.
In contrast, \emph{\textbf{ShaRP}} preserves up to 99.3\% of vanilla accuracy even with aggressive shallow-layer pruning (72\%). Its advantage is particularly pronounced on long-form video benchmarks such as LongVideoBench~\cite{wu2024longvideobench} and MLVU~\cite{zhou2024mlvu}, where it surpasses the second-best method by 2.3\% and 3.7\% at 6.03 TFLOPs, respectively. These gains arise from ShaRP’s ability to aggregate information across temporally distant frames. 
Across all retention ratios, \textit{\textbf{ShaRP}} consistently achieves competitive performance, demonstrating its robustness and adaptability.

\noindent\textbf{Cross-Backbone Generalization.}
To evaluate the cross-model generalization of our approach, we extend experiments to additional VLLM backbones, including LLaVA-Video~\cite{zhang2024video}, Qwen2-VL~\cite{wang2024qwen2}, and Qwen2.5-VL~\cite{bai2025qwen2}. These models differ substantially in visual tokenization and feature aggregation mechanisms. For example, Qwen2-VL~\cite{wang2024qwen2} employs 3D convolutions with MLP-based token aggregation, whereas LLaVA-Video~\cite{zhang2024video} adopts newline token insertion and hierarchical pooling strategies.
All methods are evaluated under identical input resolutions and comparable computational budgets. As shown in Tab.~\ref{tab:qwen}, we report results grouped by similar TFLOPs and retention ratios across MVBench~\cite{li2024mvbench,shahroudy2016ntu}, LongVideoBench~\cite{wu2024longvideobench}, and VideoMME~\cite{fu2025video}. ShaRP \textit{consistently achieves the best performance} across the evaluated settings, demonstrating robust generalization across diverse model architectures and tokenization strategies.

\input{table/compatibility}
\input{table/latency2}

\begin{figure*}[!t]
    \centering
    \includegraphics[width=1\linewidth]{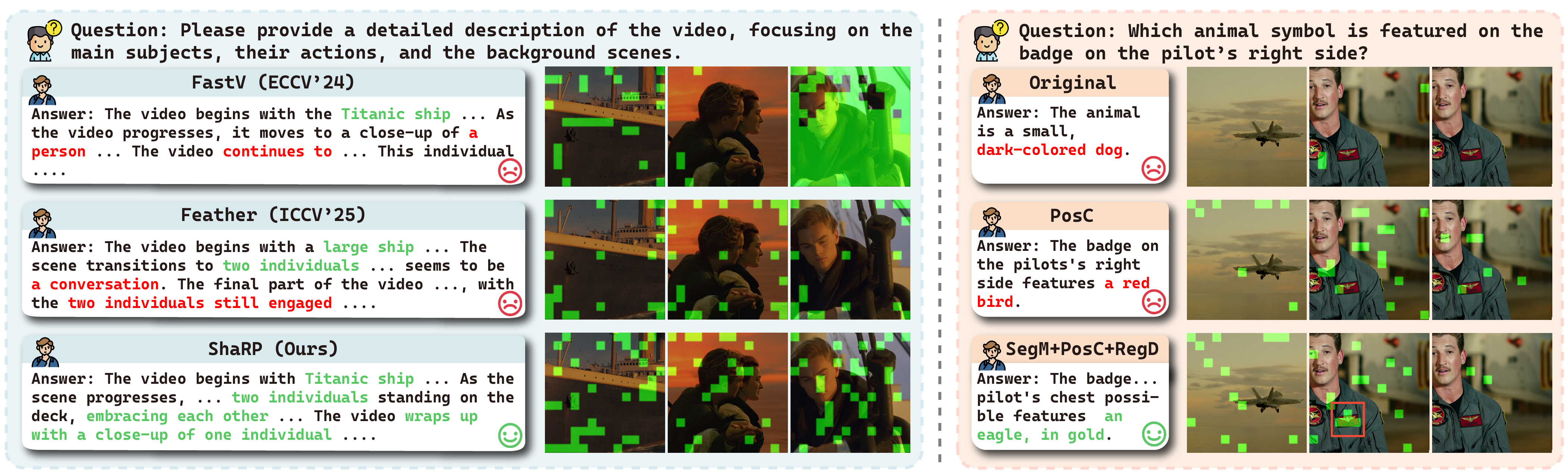} 
    \caption{\textbf{Comparison and Ablation Visualization.} 
\textbf{Left:} Comparison between \textit{ShaRP} and prior attention-based pruning methods FastV~\cite{chen2024image} and Feather~\cite{endo2025feather}. 
\textbf{Right:} Visualization of ablation results illustrating the effect of each proposed component.}
\label{fig:ablation}
\vspace{-8pt}
\end{figure*}

\noindent\textbf{Compatibility Evaluation.}
We further test ShaRP’s compatibility by integrating it into two representative pruning frameworks, \emph{VisionZip}~\cite{yang2025visionzip} and \emph{HoliTom}~\cite{shao2025holitom}, covering both outer-LLM and two-stage paradigms. As summarized in Tab.~\ref{tab:compatibility}, ShaRP consistently enhances both frameworks’ performance under equal FLOPs, confirming its compatibility.

\noindent\textbf{Inference Efficiency.}
Tab.~\ref{tab:latency} reports the inference efficiency of ShaRP on LLaVA-OneVision~\cite{lillava} and LLaVA-Video~\cite{zhang2024video}. ShaRP preserves strong performance while substantially reducing computation. For example, on LLaVA-OneVision, at 23.5\% retention, TFLOPs drop to 9.31 and the prefill latency decreases from 1181.6 ms to 354.3 ms (3.3$\times$ speedup); at 14.1\% retention, it further reduces to 233.7 ms (5.1$\times$). ShaRP also delivers up to a 4.9$\times$ prefill speedup on LLaVA-Video. 
To assess practical scalability, Fig.~\ref{fig:latency} reports latency as the input length increases. The near-linear growth pattern and limited incremental overhead demonstrate that ShaRP maintains efficient runtime behavior as video length increases.

\noindent\textbf{Visualization Comparison.}
We further perform visualization analysis in Fig.~\ref{fig:ablation} (left). As shown, \textbf{\textit{FastV}}~\cite{chen2024image} suffers from positional bias, omitting important visual segments, whereas \textbf{\textit{Feather}}~\cite{endo2025feather}’s partial debiasing and uniform sampling are less robust to scene transitions in long video QA tasks. In contrast, \textbf{\textit{ShaRP}} ensures comprehensive coverage and adapts to temporal content variations, producing semantically coherent token selection even under aggressive compression. Additional examples can be found in the supplementary material.

\subsection{Ablation Study}
\noindent\textbf{Ablation and Synergy Analysis.}
We conduct ablation experiments on our three modules—Segment-Aware Causal Masking (SegM, Sec.~\ref{sec:aggregation}), Positional Bias Calibration (PosC, Sec.~\ref{sec:debiasing}), and Register Token Deduplication (RegD, Sec.~\ref{sec:redundancy})—under a 23.5\% token retention ratio, with results in Tab.~\ref{tab:ablation}. PosC alone improves the average score from 55.8 to 57.2 by enhancing focus on critical visual regions. Adding SegM (PosC+SegM) further increases performance to 57.6, mitigating attention collapse and stabilizing temporal-contextual aggregation. Combining PosC with RegD raises the score to 57.7, demonstrating RegD’s effectiveness in compressing redundant tokens and enriching structural representation. Even without PosC, SegM+RegD reaches 57.5, confirming their independent contributions. Integrating all three modules achieves the best score of 58.1, validating the complementary benefits of each component
\input{table/abalation}

\noindent\textbf{Visualization and Analysis.}
We visualize the token selection results of the ablation study in Fig.~\ref{fig:ablation} (right) to further analyze the effect of each component.
The visualization shows that \textbf{PosC} enables the model to capture a broader and more semantically aligned visual context, while \textbf{SegM} and \textbf{RegD} encourage more diverse and informative token selection by reducing local redundancy.

%% file: table/compatibility.tex
\begin{table}[t]
  \centering
  \caption{\textbf{Compatibility Evaluation.} Results of integrating ShaRP into different pruning frameworks on LLaVA-OneVision~\cite{lillava}.}
  \setlength{\tabcolsep}{1mm}  
  \renewcommand{\arraystretch}{1.0} 
  \resizebox{\linewidth}{!}{
  \begin{tabular}{c|cc|cccc|c}
    \toprule
    \multirow{2}{*}{\makecell{\textbf{Method}}} & \multirow{2}{*}{\makecell{\textbf{Retention}\\\textbf{Ratio $\boldsymbol{R}$ ↓}}} & \multirow{2}{*}{\makecell{\textbf{TFLOPs ↓}}}
     & \textbf{MVBench ↑} & \multirow{2}{*}{\makecell{\textbf{LongVideo}\\\textbf{Bench ↑}}} & \textbf{MLVU ↑} & \textbf{VideoMME ↑} & \multirow{2}{*}{\makecell{\textbf{Avg}\\\textbf{Score ↑}}} \\
     & & & & & & & \\
    \midrule
    VisionZip~\cite{yang2025visionzip}$_{\text{CVPR'25}}$ & 12.8\% & 4.3 & 55.6 & 51.6 & 60.9 & 55.4 & 55.9 \\
    \rowcolor{blue!10} \quad + \textbf{(ShaRP)} & 20\%/12.4\% & 4.3 & \textbf{56.4} & \textbf{55.0} & \textbf{62.3} & \textbf{57.0} & \textbf{57.7} \\
    HoliTom~\cite{shao2025holitom}$_{\text{ (NeurIPS'25)}}$ & 15\%/7.5\% & 4.3 & \textbf{58.4} & 55.7 & \textbf{62.1} & 57.2 & 58.4 \\
    \rowcolor{blue!10} \quad + \textbf{(ShaRP)} & 15\%/7.5\% & 4.3 & 58.2 & \textbf{56.5} & 61.9 & \textbf{58.0} & \textbf{58.7} \\
    \bottomrule
  \end{tabular}}
  \label{tab:compatibility}
\vspace{-8pt}
\end{table}

%% file: table/latency2.tex
\setlength\tabcolsep{.4em}
\begin{table}[t]
    \begin{minipage}[t]{.54\linewidth}
    \vspace{0pt}
        \centering
        \captionof{table}{\textbf{Inference efficiency.}
“Prefill Time”: Time for model to generate first token; “Generate Time”: Time for model to generate response.}
        \resizebox{\linewidth}{!}{
            \begin{tabular}{lccc}
            \toprule
            \multirow{2}{*}{\makecell{\textbf{Retention}\\\textbf{Ratio $\boldsymbol{R}$ ↓}}} & \multicolumn{2}{c}{\textbf{Latency (ms)}} & \multirow{2}{*}{\textbf{Average ↑}} \\
            \cmidrule(l{0.5ex}r{0.5ex}){2-3}
            & \textbf{Prefill} & \textbf{Generate} & \\
            \midrule
            LLAVA-OV & 1181.6 (1.0$\times$) & 1301.3 (1.0$\times$) & 59.0 (100\%)\\

            9.31 (23.5\%) & 354.3 (3.3$\times$) & 442.2 (2.9$\times$)  & 58.1 (98.5\%) \\

            6.03 (14.1\%) & 233.7 (5.1$\times$) & 357.2 (3.6$\times$)  & 57.4 (97.2\%) \\            \midrule
            LLAVA-Video & 2404.1 (1.0$\times$) & 2796.2 (1.0$\times$) & 60.8 (100\%)\\

            16.3 (23.5\%) & 674.8 (3.6$\times$) & 883.5 (3.2$\times$)  & 59.7 (98.2\%) \\

            10.3 (14.1\%) &487.8 (4.9$\times$)  & 699.3 (4.0$\times$)  & 57.2 (94.1\%) \\
            
            \bottomrule
            \end{tabular}
        }
        \label{tab:latency}
    \end{minipage}\hfill
    \begin{minipage}[t]{.45\linewidth}
    \vspace{0pt}
        \centering
        \includegraphics[width=\linewidth,height=0.53\linewidth]{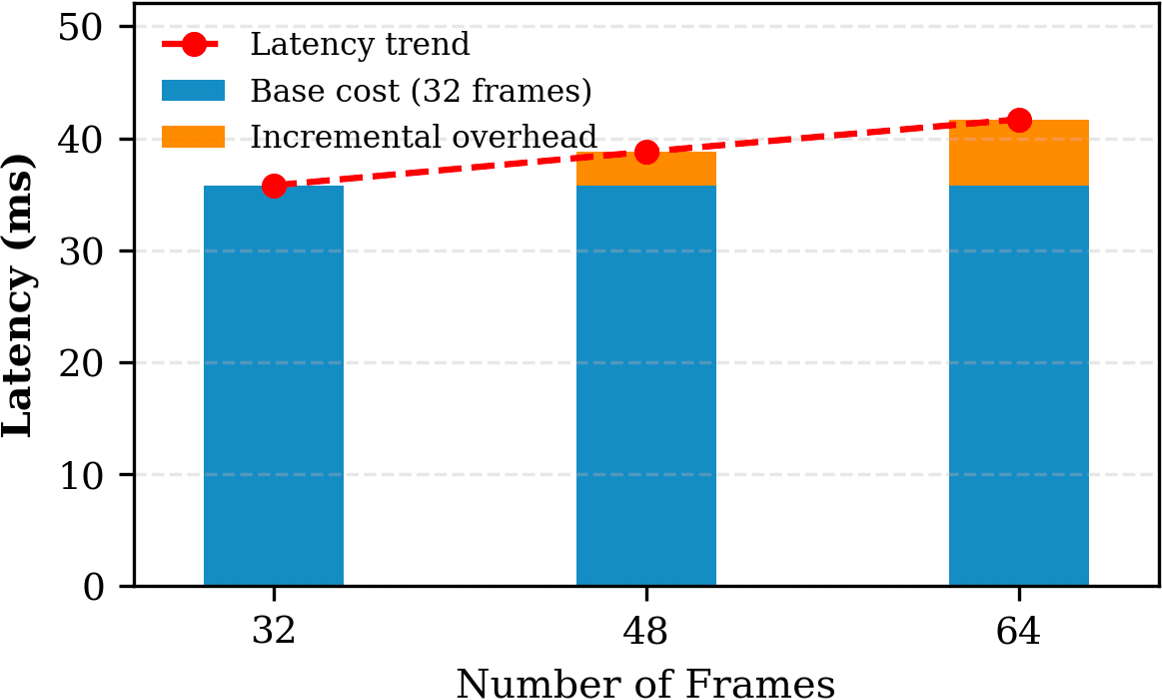}
        \captionof{figure}{\textbf{Latency scaling with increasing video length.}
Prefill latency on LLaVA-Video~\cite{zhang2024video} under ShaRP (14.1\% retention), frames 32 to 64.
}
        \label{fig:latency}
    \end{minipage}
\end{table}

%% file: table/abalation.tex
\begin{table}[t]
\centering
\caption{\textbf{Ablation study on components.} 
Results on LLaVA-OneVision~\cite{lillava}, showing the individual and combined effects of SegM, PosC, and RegD.}
\renewcommand{\arraystretch}{1}
\setlength{\tabcolsep}{1mm}

\resizebox{1\linewidth}{!}{
\begin{tabular}{c|ccccc}

\toprule
\textbf{Method} & \textbf{MVBench ↑} & \makecell{\textbf{LongVideo} \\ \textbf{Bench ↑}} & \textbf{MLVU ↑} & \textbf{VideoMME ↑} & {\makecell{\textbf{Avg}\\\textbf{Score ↑}}} \\
\midrule
Original & 54.4 & 53.0 & 61.0 & 54.7 & 55.8 \\ \midrule
PosC & 56.4 & 55.2 & 60.5 & 56.8 & 57.2 \\
PosC+SegM  & 56.6 & 55.5 & 60.9 & 57.2 & 57.6 \\
PosC+RegD & 56.7 & 55.6 & 61.5 & 57.0 & 57.7 \\
SegM+RegD & 56.4 & 54.7 & 61.8 & 57.1 & 57.5 \\
PosC+SegM+RegD (ShaRP) &\bluetext{57.1} & \bluetext{55.8} & \bluetext{61.9} & \bluetext{57.6} & \bluetext{58.1} \\
\bottomrule
\end{tabular}
}
\vspace{-8pt}
\label{tab:ablation}
\end{table}

%% file: sec/5_conclusion.tex
\section{Conclusion}
\label{sec:conclusion}
In this work, we propose ShaRP, a training-free framework addressing the critical challenge of shallow-layer pruning in Video Large Language Models (VLLMs). We identify three key factors causing attention-based pruning to fail in early layers: attention collapse, positional encoding bias, and register token redundancy. To tackle these limitations in shallow-layer attention, we introduce a synergistic three-stage pipeline that enhances feature aggregation, calibrates attention distributions and removes redundancy. Extensive experiments show that our approach maintains strong cross-modal understanding while substantially reducing computational overhead and exhibiting broad compatibility, paving the way for efficient and scalable VLLM inference.